%% file: manuscript.tex
\documentclass{article}
\usepackage[preprint]{log_2022}
\usepackage{amsmath, amssymb,amsthm}
\usepackage{graphicx,subfigure}
\usepackage{color}
\usepackage{url}
\usepackage[a]{esvect}
\usepackage{enumitem}
\usepackage[utf8]{inputenc}
\usepackage{setspace}
\usepackage{wrapfig}
\usepackage{mdframed}
\usepackage{float}
\usepackage{xcolor}
\usepackage{booktabs}           
\usepackage{multirow}           
\usepackage{amsfonts}           

\usepackage{titlesec}

\usepackage[numbers,compress,sort]{natbib}
\usepackage[colorinlistoftodos]{todonotes}  
\usepackage[backref=page]{hyperref} 
\usepackage{cleveref}

\def\doubleblind{1}
\if\doubleblind1
\newcommand{\blinded}[1]{{\color{red} [blinded]}} 
\else 
\newcommand{\blinded}[1]{#1}
\fi

\titlespacing*{\section}
  {0pt}{.3\baselineskip}{.15\baselineskip}
\titlespacing*{\paragraph}
  {0pt}{.2\baselineskip}{.2\baselineskip}

\author[Lisi Qarkaxhija et al.]{%
Lisi Qarkaxhija\\
\institute{Chair of Machine Learning for Complex Networks}\\
\institute{Center for Artificial Intelligence and Data Science (CAIDAS)}\\
\institute{Julius-Maximilians-Universit\"at W\"urzburg, DE}\\
\email{lisi.qarkaxhija@uni-wuerzburg.de}\And
 Vincenzo Perri\\
\institute{Data Analytics Group}\\
\institute{Department of Informatics}\\
\institute{University of Zurich, CH}\\
\email{perri@ifi.uzh.ch} \And
 Ingo Scholtes\thanks{also with Data Analytics Group, Department of Informatics, University of Zurich, Zurich, CH}\\
\institute{Chair of Machine Learning for Complex Networks}\\
\institute{Center for Artificial Intelligence and Data Science (CAIDAS)}\\
\institute{Julius-Maximilians-Universit\"at W\"urzburg, DE}\\
\email{ingo.scholtes@uni-wuerzburg.de}
}

\newtheorem{definition}{Definition}

\begin{document}

\title{De Bruijn goes Neural: Causality-Aware Graph Neural Networks for Time Series Data on Dynamic Graphs}

\maketitle

\begin{abstract}
We introduce De Bruijn Graph Neural Networks (DBGNNs), a novel time-aware graph neural network architecture for time-resolved data on dynamic graphs.
Our approach accounts for temporal-topological patterns that unfold in the causal topology of dynamic graphs, which is determined by \emph{causal walks}, i.e. temporally ordered sequences of links by which nodes can influence each other over time.
Our architecture builds on multiple layers of higher-order De Bruijn graphs, an iterative line graph construction where nodes in a De Bruijn graph of order $k$ represent walks of length $k-1$, while edges represent walks of length $k$.

We develop a graph neural network architecture that utilizes De Bruijn graphs to implement a message passing scheme that follows a non-Markovian dynamics, which enables us to learn patterns in the causal topology of a dynamic graph.
Addressing the issue that De Bruijn graphs with different orders $k$ can be used to model the same data set, we further apply statistical model selection to determine the optimal graph topology to be used for message passing.
An evaluation in synthetic and empirical data sets suggests that DBGNNs can leverage temporal patterns in dynamic graphs, which substantially improves the performance in a supervised node classification task.

\end{abstract}

\section{Introduction} 
\label{sec:intro}

Graph Neural Networks (GNNs) \cite{Hamilton2020_GraphRepresentationLearning,Wu2021_GNNSurvey} have become a cornerstone for the application of deep learning to data with a non-Euclidean, relational structure.
Different flavors of GNNs have been shown to be highly efficient for tasks like node classification, representation learning, link prediction, cluster detection, or graph classification.

The popularity of GNNs is largely due to the abundance of data that can be represented as graphs, i.e. as a set of \emph{nodes} with pairwise connections represented as \emph{links}.
However, we increasingly have access to \emph{time-resolved data} that not only capture which nodes are connected to each other, but also when and in which temporal order those connections occur.
A number of works in computer science, network science, and interdisciplinary physics have highlighted how the \emph{temporal dimension} of dynamic graphs, i.e. the timing and ordering of links, influences the \emph{causal topology} of networked systems, i.e. which nodes can possibly influence each other over time \cite{Kempe2000_Temporal,Holme2012,BadieModiri2020}.
In a nutshell, if an undirected link $(a,b)$ between two nodes $a$ and $b$ occurs \emph{before} an undirected link $(b,c)$, node $a$ can causally influence node $c$ via node $b$.
If the temporal ordering of those two links is reversed, node $a$ cannot  influence node $c$ via $b$ due to the directionality of the arrow of time. 
This simple example shows that the arrow of time in dynamic graphs limits possible causal influences between nodes beyond what we would expect based on the mere topology of links.

Beyond such toy examples, a number of recent studies in network science, computer science, and interdisciplinary physics have shown that the temporal ordering of links in real time series data on graphs has non-trivial consequences for the properties of networked systems, e.g. for reachability and percolation \cite{Lentz2013,BadieModiri2022}, diffusion and epidemic spreading \cite{,Pfitzner2013_prl,Scholtes2014_natcomm}, node rankings and community structures \cite{Rosvall2014}. 
It had further been shown that this interesting aspect of dynamic graphs can be understood using a variant of \emph{De Bruijn graphs} \cite{DeBruijn1946}, i.e. static higher-order graphical models \cite{Scholtes2017,Lambiotte2019,Scholtes2014_natcomm} of causal paths that capture both the temporal and the topological dimension of time series data on graphs.

While the generalization of network analysis techniques like node centrality measures 
and community detection \cite{Rosvall2014,Scholtes2017}, or graph embedding \cite{Belth2019_evo} to such higher-order models has been successful, to the best of our knowledge no generalizations of Graph Neural Networks to higher-order De Bruijn graphs have been proposed \cite{eliassirad2021,Krieg2022_dge}.
Such a generalization bears several promises:
First it could enable us to apply well-known and efficient gradient-based learning techniques in a static neural network architecture that is able to learn patterns in the causal topology of dynamic graphs that are due to the temporal ordering of links.
Second, making the temporal ordering of links in time-stamped data a first-class citizen of graph neural networks, this generalization could also be an interesting approach to incorporate a necessary condition for \emph{causality} into state-of-the-art geometric deep learning techniques, which often lack meaningful ways to represent time.
Finally, a combination of higher-order De Bruijn graph models with graph neural networks enable us to apply frequentist and Bayesian techniques to learn the ``optimal'' order of a De Bruijn graph model for a given time series, providing new ways to combine statistical learning and model selection with graph neural networks.

Addressing this gap, our work generalizes graph neural networks to high-dimensional De Bruijn graph models for causal paths in time-stamped data on dynamic graphs. 
We obtain a novel causality-aware graph neural network architecture for time series data that makes the following contributions:
\begin{itemize}[noitemsep,topsep=0pt]
\item We develop a graph neural network architecture that generalizes message passing to multiple layers of higher-order De Bruijn graphs. The resulting De Bruijn Graph Neural Network (DBGNN) architecture leads to a \emph{non-Markovian} message passing, whose dynamics matches correlations in the temporal ordering of links, thus enabling us to learn patterns that shape the causal topology of dynamic graphs.
\item We evaluate our proposed architecture both in empirical and synthetically generated dynamic graphs and compare its performance to graph neural networks as well as (time-aware) graph representation learning techniques. We find that our method yields superior node classification performance.
\item We combine this architecture with statistical model selection to infer the optimal higher order of a De Bruijn graph. This yields a two-step learning process, where (i) we first learn a parsimonious De Bruijn graph model that neither under- nor overfits patterns in a dynamic graph, and (ii) we apply message passing and gradient-based optimization to the inferred graph in order to address graph learning tasks like node classification or representation learning.
\end{itemize}

Our work builds on the --to the best of our knowledge-- novel combination of (i) statistical model selection to infer optimal higher-order graphical models for causal paths in dynamic graphs, and (ii) gradient-based learning in a neural network architecture that uses the inferred higher-order graphical models as message passing layers.
Thanks to this approach, our architecture performs message passing in an optimal graph model for the causal paths in a given dynamic graph.
The results of our evaluation confirm that this explicit regularization of the message passing layers enables us to considerably improve performance in a node classification task.
The remainder of this paper is structured as follows: In \cref{sec:background} we introduce the background of our work and formally state the problem that we address, in \cref{sec:architecture} we introduce the De Bruijn graph neural network architecture, in \cref{sec:experiments} we experimentally validate our method in synthetic and empirical data on dynamic graphs, and in \cref{sec:conclusion} we summarize our contributions and highlight opportunities for future research.
We have implemented our architecture based on the graph learning library \texttt{pyTorch Geometric} \cite{fey2019fast} and release the code of our experiments as an Open Source package.

\section{Background and Problem Statement} 
\label{sec:background}

\paragraph{Basic definitions} We consider a dynamic graph $G^{\mathcal{T}}=(V, E^{\mathcal{T}})$ with a (static) set of nodes $V$ and time-stamped (directed) edges $(v,w;t) \in E^{\mathcal{T}} \subseteq V \times V \times \mathbb{N}$ where --without loss of generality-- integer timestamps $t$ represent the instantaneous time at which a pair of nodes $v, w$ is connected \cite{Holme2012}.
While many real-world network data exhibit such timestamps, for the application of graph neural networks we often consider a \emph{time-aggregated projection} $G(V,E)$ along the time axis, where a (static) edge $(v,w) \in E$ exists iff $\exists t \in \mathbb{N}: (v,w) \in E^{\mathcal{T}}$.
We can further consider edge weights $w: E \rightarrow \mathbb{N}$ defined as $w(v,w) := |\{t \in \mathbb{N}: (v,w;t) \in E^{\mathcal{T}} \}|$, i.e. we use $w(v,w)$ to count the number of temporal activations of $(v,w)$.

A key motivation for the study of graphs as models for complex systems is that --apart from \emph{direct} interactions captured by edges $(v,w)$-- they facilitate the study of \emph{indirect} interactions between nodes via \emph{paths} or \emph{walks} in a graph.
Formally, we define a walk $v_0, v_1, \ldots, v_{l-1}$ of length $l$ in a graph $G=(V,E)$ as any sequence of nodes $v_i \in V$ such that $(v_{i-1}, v_{i}) \in E$ for $i=1, \ldots, l-1$.
The length $l$ of a walk captures the number of traversed edges, i.e. each node $v \in V$ is a walk of length zero, while each edge $(v,w)$ is a walk of length one.
We further call a walk $v_0, v_1, \ldots, v_{l-1}$ a \emph{path} of length $l$ from $v_0$ to $v_{l-1}$ iff $v_i \neq v_j$ for $i\neq j$, i.e. a path is a walk between a set of distinct nodes.

\paragraph{Causal walks and paths in dynamic graphs} In a static graph $G=(V,E)$, the topology--i.e. which nodes can \emph{directly and indirectly} influence each other via edges, walks, or paths-- is completely determined by the edges $E$.
This is is different for dynamic graphs, which can be understood by extending the definition of walks and paths to \emph{causal concepts} that respect the \emph{arrow of time}:
\begin{definition}
\label{def:causalwalk}
For a dynamic graph $G^{\mathcal{T}}=(V,E^{\mathcal{T}})$, we call a node sequence $v_0, v_1, \ldots, v_{l-1}$ a \emph{causal walk} iff the following two conditions hold: (i) $(v_{i-1}, v_{i}; t_i) \in E^{\mathcal{T}}$ for $i=1, \ldots, l-1$ and (ii) $0 < t_j - t_i \leq \delta$ for $i<j$ and some $\delta > 0$.
\end{definition}
The first condition ensures that nodes in a dynamic graph can only indirectly influence each other via a causal walk iff a corresponding walk exists in the time-aggregated graph.
Due to $0 < t_j - t_i$ for $i<j$, the second condition ensures that time-stamped edges in a causal walk occur in the correct chronological order, i.e. timestamps are monotonically increasing \cite{Kempe2000_Temporal,Holme2012}.
As an example, two time-stamped edges $(a,b;1), (b,c;2)$ constitute a causal walk by which information from node $a$ starting at time $t_1=1$ can reach node $c$ at time $t_2=2$ via node $b$, while the same edges in reverse temporal order $(a,b;2), (b,c;1)$ do not constitute a causal walk.
While this definition of a causal walk does not impose an \emph{upper bound} on the time difference between consecutive time-stamped edges constituting a causal walk, it is often reasonable to define a time limit $\delta > 0$, i.e. a time difference beyond which consecutive edges are not considered to contribute to a causal walk.
As an example, two time-stamped edges $(a,b;1), (b,c;100)$ constitute a causal walk by which information from node $a$ starting at time $t_1=1$ can reach node $c$ at time $t_2=100$ via node $b$ for $\delta=150$, while they do not constitute a causal walk for $\delta=5$.
This time-limited notion of causal or time-respecting walks is characteristic for many real networked systems in which processes or agents have a finite time scale or ``memory'', which rules out infinitely long gaps between consecutive causal interactions \cite{Holme2012,BadieModiri2020}.
Analogous to the definition in a static network, we finally define a \emph{causal path} $v_0, v_1, \ldots, v_{l-1}$ of length $l$ from node $v_0$ to node $v_{l-1}$ as a causal walk with $v_i \neq v_j$ for $i \neq j$.

\paragraph{Non-Markovian characteristics of dynamic graphs} The above definition of causal walks and paths in dynamic graphs has important consequences for our understanding of the \emph{topology} of dynamic graphs, i.e. which nodes can directly and indirectly influence each other directly via walks or paths.
Moreover, it has important consequences for graph learning and network analysis tasks such as node ranking, cluster detection, or embedding\cite{Rosvall2014,Scholtes2014_natcomm,salnikov2016using,Scholtes2017,Lambiotte2019}.
This additional complexity of dynamic graphs is due to the fact that the topology of a static graph $G=(V,E)$ can be fully understood based on the \emph{transitive hull of edges}, i.e. the presence of two edges $(u,v) \in E$ and $(v,w) \in E$ implies that nodes $u$ and $w$ can indirectly influence each other via a walk or path, which we denote as $u \rightarrow^* w$.
This not only enables us to use standard algorithms, e.g. to calculate (shortest) paths, it also implies that we can use matrix powers, eigenvalues and eigenvectors to analyze topological properties of a graph.
In contrast, in dynamic graphs the chronological order of time-stamped edges can break transitivity, i.e. $(u,v;t) \in E$ and $(v,w;t') \in E$ does \emph{not} necessarily imply $u \rightarrow^* w$, which invalidates graph analytic approaches \cite{Lambiotte2019}.

To study the question how correlations in the temporal ordering of time-stamped edges influence the \emph{causal topology} of a dynamic graph, we can take a statistical modelling perspective.
We can, for instance, consider causal walks as sequences of random variables that can be modelled via a Markov chain of order $k$ over a discrete state space $V$ \cite{Scholtes2017}.
In other words, we model the sequence of nodes $v_0, \ldots, v_{l-1}$ on causal walks as 
$P(v_i|v_{i-k}, \ldots, v_{i-1})$
where $k-1$ is the length of the ``memory'' of the Markov chain. 
For $k=1$ we have a memoryless, first-order Markov chain model $P(v_i|v_{i-1})$, where the next node on the walk exclusively depends on the current node. 
From the perspective of dynamic graphs with time-stamped link sequences, this corresponds to a case where the causal walks of the dynamic graph are exclusively determined by the topology (and possibly frequency) of edges, i.e. there are no correlations in the temporal ordering of time-stamped edges and the causal topology of the dynamic matches the topology of the corresponding time-aggregated graph.
If the need a Markov order $k>1$, the sequence of nodes traversed by causal walks exhibits \emph{memory}, i.e. the next node on a walk not only depends on the current one but also on the history of past interactions. 
The presence of such higher-order correlations in dynamic graphs is associated with more complex causal topologies that (i) cannot be reduced to the topology of the associated time-aggregated network, and (ii) have interesting implications for spreading and diffusion processes and spectral properties \cite{Scholtes2014_natcomm}, node centralities \cite{Scholtes2017}, and community structures \cite{Rosvall2014}.

\paragraph{Higher-order De Bruijn graph models of causal topologies} 
The use of higher-order Markov chain models for causal paths leads to an interesting novel view on the relationship between graph models and time series data on dynamic graphs.
In this view, the common (weighted) time-aggregated graph representation of time-stamped edges corresponds to a \emph{first-order graphical model}, where edge weights capture the statistics of edges, i.e. causal paths of length one. 
A normalization of edge weights in this graph yields a first-order Markov model of causal walks in a dynamic graph.
Similarly, a graphical representation of higher-order Markov chain model of causal walks  can be used to capture non-Markovian patterns in the temporal sequence of time-stamped edges.
However, different from higher-order Markov chain models of general categorical sequences, a higher-order model of causal paths in dynamic graphs must account for the fact that the set of possible \emph{causal paths} is constrained by the topology of the corresponding static graph (i.e. condition (i) in \Cref{def:causalwalk}).
To account for this we define a higher-order De Bruijn graph model of causal walks~\cite{DeBruijn1946}:

\begin{definition}[\bf \(k\)-th order De Bruijn graph model]
\label{def:debruijn}
For a dynamic graph \(G^{\mathcal{T}}=(V,E^{\mathcal{T}})\) and $k \in \mathbb{N}$, a \(k\)-th order De Bruijn graph model of causal paths in $G^{\mathcal{T}}$ is a graph $G^{(k)}=(V^{(k)}, E^{(k)})$, with \(u:=(u_0, u_1, \ldots, u_{k-1}) \in V^{(k)}\) a causal walk of length \(k-1\) in \(G^{\mathcal{T}}\) and \((u,v) \in E^{(k)}\) iff (i) \(v=(v_1, \ldots, v_{k})\) with \(u_{i} = v_{i}\) for \(i=1, \ldots, k-1\) and (ii) \(u \oplus v = (u_0, \ldots, u_{k-1}, v_{k})\) a causal walk of length \(k\) in \(G^{\mathcal{T}}\).
\end{definition}
We note that any two adjacency nodes $u, v \in V^{k}$ in a \(k\)-th order De Bruijn graph $G^{(k)}$ represent two causal walks of length \(k-1\) that overlap in exactly \(k-1\) nodes, i.e. each edge $(u,v)\in E^{(k)}$ represents a causal walk of length \(k\).
We can further use edge weights $w: E^{(k)} \rightarrow \mathbb{N}$ to capture the frequencies of causal paths of length $k$.
The (weighted) time-aggregated graph \(G\) of a dynamic graph trivially corresponds to a first-order De Bruijn graph, where (i) nodes are causal walks of length zero and (ii) edges $E=E^{(1)}$ capture causal walks of length one (i.e. edges) in \(G^{\mathcal{T}}\).
To construct a second-order De Bruijn graph $G^{(2)}$ we can perform a line graph transformation of a static graph $G=G^{(1)}$, where each edge $(u_0,u_1), (u_1, u_2) \in E^{(2)}$ captures a causally ordered sequence of two edges $(u_0, u_1; t)$ and $(u_1, u_2; t')$.
A $k$-th order De Bruijn graph can be constructed by a repeated line graph transformation of a static graph $G$.
Hence, De Bruijn graphs can be viewed as generalization of common graph models to a higher-order, static graphical model of causal walks of length $k$, where walks of length $l$ in $G^k$ model causal walks of length $k+l-1$ in $G^{\mathcal{T}}$ \cite{Scholtes2014_natcomm,Lambiotte2019}.

\begin{figure}[!htb]
\begin{center}
\includegraphics[width=\textwidth]{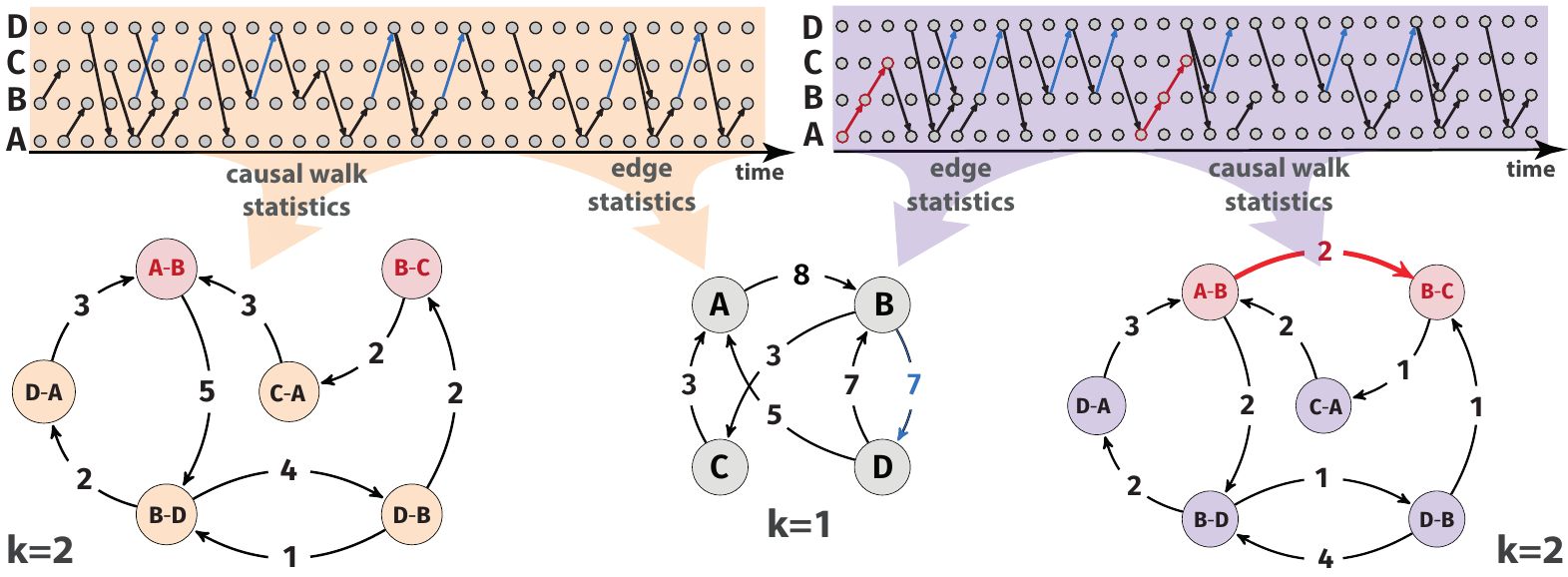}
\end{center}

\caption{Simple example for two dynamic graphs with four nodes and $33$ directed time-stamped edges (top left and right). The two graphs only differ in terms of the temporal ordering of edges. Frequency and topology of edges are identical, i.e. they have the same first-order time-aggregated weighted graph representation (center).
Due to the arrow of time, causal walks and paths differ in the two dynamic graphs: Assuming $\delta=1$, in the left dynamic graph node $A$ cannot causally influence $C$ via $B$, while such a causal path is possible in the right graph. 
A second-order De Bruijn graph representation of causal walks in the two graphs (bottom left and right) captures this difference in the causal topology. Building on such causality-aware graphical models, in our work we define a graph neural network architecture that is able to learn patterns in the causal topology of dynamic graphs.}
\label{fig:network_model}
\end{figure}

De Bruijn graphs have interesting mathematical properties that connect them to trajectories of subshifts of finite type as well as to dynamical systems and ergodic theory \cite{ChungDeBruijn1992}.
For the purpose of our work, they provide the advantage that we can use $k$-th order De Bruijn graphs to model the \emph{causal topology} in dynamic graphs.
We illustrate this in \cref{fig:network_model}, which shows two dynamic graphs with four nodes and 33 time-stamped links. 
These dynamics groups only differ in term of the temporal ordering of edges, i.e. they have the same (first-order) weighted time-aggregated graph representation (center).
Moreover, this first-order representation wrongly suggests that node $A$ can influence node $C$ by a path via node $B$.
While this is true in the dynamic graph on the right (see red causal paths), no corresponding causal path from $A$ via $B$ to $C$ exists in the dynamic graph on the left.
A second-order De Bruijn graph model (bottom left and right) captures the fact that the causal path from $A$ via $B$ to $C$ is absent in the right example.
This shows that, different from commonly used static graph representations, the edges of a $k$-th order De Bruijn graph with $k>1$ are sensitive to the temporal ordering of time-stamped edges.
Hence, static higher-order De Bruijn graphs can be used to model the causal topology in a dynamic graph.
We can view a $k$-th order De Bruijn graph in analogy to a $k$-th order Markov model, where a directed link from node $(u_0, \ldots, u_{k-1})$ to node $u=(u_1, \ldots, u_{k}, u_k)$ captures a walk from node $u_k$ to $u_{k+1}$ in the underlying graph, with a memory of $k$ previously visited nodes $u_0, \ldots, u_{k-1}$.
This approach has been used to analyze how the causal topology of dynamic graphs influences node ranking in dynamic graphs \cite{Rosvall2014,Scholtes2017}, the modelling of random walks and diffusion \cite{Scholtes2014_natcomm}, community detection \cite{salnikov2016using,Rosvall2014}, time-aware static graph 
embedding \cite{Saebi2020_HONEM,Belth2019_evo}.

Moreover, several works have proposed heuristic, frequentist and Bayesian methods to infer the optimal order of higher-order graph models of causal paths given time series data on dynamic graphs \cite{Rosvall2014,Scholtes2017,Xu2016,Petrovic2022_WWW}.

\paragraph{Problem Statement and Research Gap} 
The works above provide the background for the generalization of graph neural networks to higher-order De Bruijn graph models of causal walks in dynamic graphs, which we propose in the following section.
Following the terminology in the network science community, higher-order De Bruijn graph models can be seen as one particular type of \emph{higher-order network models} \cite{Lambiotte2019,Torres2021,benson2021higher}, which capture (causally-ordered) sequences of interactions between more than two nodes, rather than dyadic edges.
They complement other types of popular higher-order network models (like, e.g. hypergraphs, simplicial complexes, or motif-based adjacency matrices) that consider (unordered) non-dyadic interactions in static networks, and which have been used to generalize graph neural networks to non-dyadic interactions \cite{Feng2018,Huang2021}.

To the best of our knowledge, De Bruijn graph models have not been combined with recent advanced in graph neural networks.

Closing this gap, we propose a causality-aware graph convolutional network architecture that uses an \emph{augmented} message passing scheme \cite{velivckovic2022message} in higher-order De Bruijn graphs to capture patterns in the causal topology of dynamic graphs.

\section{De Bruijn Graph Neural Network Architecture} 
\label{sec:architecture}
We now introduce the De Bruijn Graph Neural Network (DBGNN) architecture with an \emph{augmented message passing} \cite{velivckovic2022message} scheme whose dynamics matches the non-Markovian characteristics of dynamic graphs, which is the key contribution of our work.
While we build on the message passing proposed for Graph Convolutional Networks (GCN) \cite{Kipf2016_GCN}, it is easy to generalize our architecture to other message passing schemes.
Our approach is based on the following three steps, which yield an easy to implement and scalable class of graph neural networks for time series and sequential data on graphs: We first use time series data on dynamic graphs to calculate statistics of causal walks of different lengths $k$. 
We use these statistics to select an higher-order De Bruijn graph model for the causal topology of a dynamic graph.
This step is parameter-free, i.e. we can use statistical learning techniques to infer an optimal graph model for the causal topology directly from time series data, without need for hyperparameter tuning or cross-validation.
We now define a graph convolutional network that builds on neural message passing in the higher-order De Bruijn graphs inferred in step one.
The hidden layers of the resulting graph convolutional network yield meaningful latent representations of patterns in the \emph{causal topology of a dynamic graph}.
Since the nodes in a $k$-th order De Bruijn graph model correspond to walks (i.e. sequences) of nodes of length $k-1$, we implement an additional bipartite layer that maps the latent space representations of sequences to nodes in the original graph.
In the following, we provide a detailed description of the three steps outlined above:

\paragraph{Inference of Optimal Higher-Order De Bruijn Graph Model}

The first step in the DBGNN architecture is the inference of the higher-order De Bruijn graph model for the causal topology in a given dynamic graph data set.
For this, we use \Cref{def:causalwalk} to calculate the statistic of causal walks of different lengths $k$ for a given maximum time difference $\delta$.
We note that this can be achieved using efficient window-based algorithms \cite{badie2020efficient,Petrovic2021}.
The statistics of causal walks in the dynamic graph allows us to apply the model selection technique proposed in \cite{Scholtes2017}, which yields the optimal higher-order of a De Bruijn graph model given the statistics of causal walks (or paths). 
The resulting (static) higher-order De Bruijn graph model is the basis for our extension of the message passing scheme for a dynamic graph with non-Markovian characteristics.

\paragraph{Message passing in higher-order De Bruijn graphs}
Standard message passing algorithms in graph neural networks use the topology of a graph to propagate (and smooth) features across nodes, thus generating hidden features that incorporate patterns in the topology of a graph.

To additionally incorporate patterns in the \emph{causal topology of a dynamic graph} we perform message passing in multiple layers of higher-order De Bruijn graphs.

Assuming a $k$-th order De Bruijn graph model $G^{(k)}=(V^{(k)}, E^{(k)})$ as defined in \Cref{def:debruijn}, the input to the first layer $l=0$ is a set of $k$-th order node features $\mathbf{h^{k,0}} = \{ \Vec{h}^{k,0}_1, \Vec{h}^{k,0}_2, \dots, \Vec{h}^{k,0}_N\}$, for $\Vec{h}^{k,0}_i \in \mathbb{R}^{H^{0}}$, where $N=|V^{(k)}|$ and $H^{0}$ is the dimensionality of initial node features.
The De Bruijn graph message passing layer uses the causal topology to learn a new set of hidden representations for higher-nodes $\mathbf{h^{k,1}} = \{ \Vec{h}^{k,1}_1, \Vec{h}^{k,1}_2, \dots, \Vec{h}^{k,1}_N\}$, with $\Vec{h}^{k,1}_i \in \mathbb{R}^{H^{1}}$  for each $k-th$ order node $i$ (corresponding to a causal walk of length $k-1$).
For layer $l$, we define the update rule of the message passing as:
\begin{equation}
\Vec{h}_v^{k,l} = \sigma \Bigg( \pmb{W}^{k,l} \sum_{\{ u \in V^{(k)}: (u,v) \in E^{(k)} \} \cup \{ v\}} \frac{w(u,v) \cdot \Vec{h}_v^{k,l-1}}{\sqrt{S(v) \cdot S(u)}}\Bigg), 
\end{equation}
where $\Vec{h}_u^{k,l-1}$ is the previous hidden representation of node $u \in V^k$, $w(u,v)$ is the weight of edge $(u,v) \in E^k$ (capturing the frequency of the corresponding causal walk as explained in \cref{sec:background}), $\pmb{W}^{k,l} \in \mathbb{R}^{H^{l} \times H^{l-1}}$ are trainable weight matrices, $S(v):=\sum_{u \in V^{(k)}} w(u,w)$ is the sum of weights of incoming edges of nodes, and $\sigma$ is a non-linear activation function.

Since the message passing is performed on a higher-order De Bruijn graph, we obtain a \emph{non-Markovian} (or rather higher-order Markovian) message passing dynamics, i.e. we perform a Laplacian smoothing that follows the non-Markovian patterns in the causal walks in the underlying dynamic graph.
Different from standard, static graph neural networks that ignore the temporal dimension of dynamic graphs, this enables our architecture to incorporate temporal patters that shape the causal topology, i.e. which nodes in a dynamic graph can influence each other directly and indirectly based on the temporal ordering of time-stamped edges (and the arrow of time).

\paragraph{First-order message passing and bipartite projection layer}
While the (static) topology of edges influences the (possible) causal walks and thus the edges in the $k$-th order De Bruin graph, it is important to note that --due to the fact that it operates on nodes $V^{(k)}$ in the \emph{higher-order} graph-- the message passing outlined above does not allow us to incorporate information on the first-order topology.
To address this issue, we additionally include message passing in the (static) time-aggregated weighted graph $G$, which can be done in parallel to the message passing in the higher-order De Bruijn graph.
The $g$ layers of this first-order message passing (whose formal definition we omit as it simply uses the GCN update rule \cite{Kipf2016_GCN}) generate hidden representations $\Vec{h}_v^{1,g}$ of nodes $v \in V$.
This approach enables us to incorporate optional node features $\Vec{h}_v^{0,g}$ (or alternatively use a one-hot-encoding of nodes).

Since the message passing in a higher-order De Bruijn graph generates hidden features for higher-order nodes $V^{(k)}$ (i.e. sequences of $k$ nodes) rather than nodes $V$ in the original dynamic graph, we finally define a bipartite graph $G^b=(V^{(k)} \cup V, E^b \subseteq V^{(k)} \times V)$ that maps node features of higher-nodes to the first-order node space.
For a given node $v \in V$, this bipartite layer sums the hidden representations $\vec{h}^{k,l}_u$ of each higher-order node $u = (u_0, \ldots, u_{k-1}) \in V^{(k)}$ with $u_{k-1} = v$ to the representation $h^{1,g}_v \in \mathbb{R}^{F^{g}}$ generated by the last layer of the first-order message passing.
Notice that the dimensions of representations in the last layers of the $k$-th and first-order message passing should satisfy $F^{g} = H^{l}$ to enable the summing of the representations.
We obtain representations $\{ \Vec{h}_u^{k,l} + \Vec{h}_v^{1,g}: \text{for } u \in V^k \text{ with }(u,v) \in E^b\}$ that are the higher-order node representations augmented by the corresponding first order representations. 
We then use a function $\mathcal{F}$ to aggregate the augmented higher-order representations at the level of first-order nodes.
In our experiments, we learn first-order node representations $h^{1,g}$ using GCN message passing with $g$ layers, allowing to integrate information on the static and the causal topology of a dynamic graph.
Formally, we define the bipartite layer as
\begin{equation}
\Vec{h}_v^b = \sigma \Bigg( \pmb{W}^b \mathcal{F}\left( \{ \Vec{h}_u^{k,l} + \Vec{h}_v^{1,g}: \text{for } u \in V^{(k)} \text{ with }(u,v) \in E^b\} \right) \Bigg),
\end{equation}
where $\Vec{h}_v^b$ is the output of the bipartite layer for node $v \in V$, and $\mathbf{W}^{b} \in \mathbb{R}^{F^{g} \times H^{l}}$ is a learnable weight matrix.
The function $\mathcal{F}$ can be SUM, MEAN, MAX, MIN.

\begin{figure}[t]
    \centering
    \includegraphics[height=5cm]{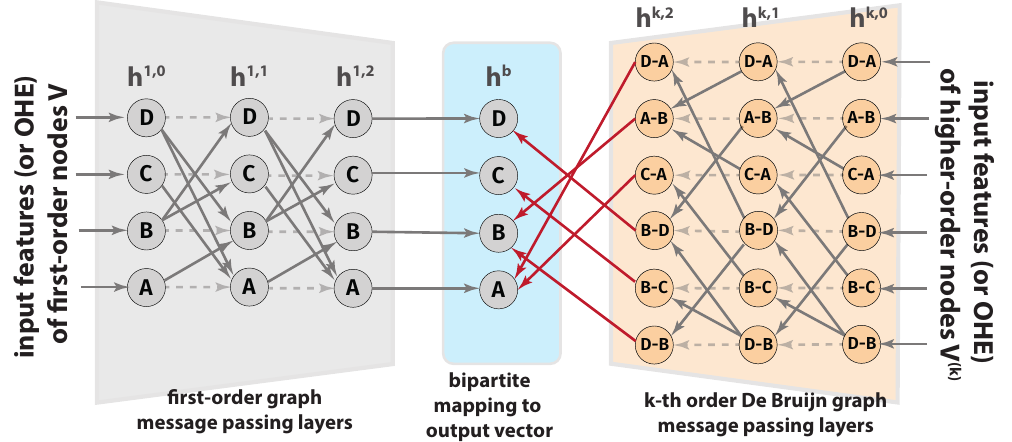}
    \caption{Illustration of DBGNN architecture with two message passing layers in first- (left, gray) and second-order De Bruijn graph (right, orange) corresponding to the dynamic graph in \Cref{fig:network_model} (left).
    Red edges represent indicate the bipartite mapping $G^b$ of higher-order node representations to first-order representations. An additional linear layer (not shown) is used for node classification.
    }
    \label{fig:bipartite_mapping}
\end{figure}

\Cref{fig:bipartite_mapping} gives an overview of the proposed neural network architecture for the dynamic graph (and associated second-order De Bruijn graph model) shown in \Cref{fig:network_model} (left).
The higher-order message passing layers on the right use the topology of the second-order De Bruijn graph in \Cref{fig:network_model} (left), while the first-order message passing layers (left) use the topology of the first-order graph.
Note that the first-order and higher-order message passing can be performed in parallel, and that the number of message passing layers do not necessarily need to be the same.
Red edges indicate the propagation of higher-order node representations to first order nodes performed in the final bipartite layer.
Due to space constraints, in \Cref{fig:bipartite_mapping} we omit the final linear layer used for classification.

\section{Experimental Evaluation}
\label{sec:experiments}

In the following, we experimentally evaluate our proposed causality-aware graph neural network architecture both in synthetic and empirical time series data on dynamic graphs. 
With our evaluation, we want to answer the following questions:

\begin{enumerate}[noitemsep,topsep=0pt]
\item[\bfseries Q1] How does the performance of De Bruijn Graph Neural Networks compare to temporal and non-temporal graph learning techniqes?
\item[\bfseries Q2] Can we use De Bruijn Graph Neural Networks to learn interpretable \emph{static} latent space representations of nodes in \emph{dynamic} graphs?

\end{enumerate}

To address those questions, we use six time series data sets on dynamic graphs that provide meta-information on node classes.
The overall statistics of the data sets can be found in \cref{table: datasetinfo},
\textbf{temp-clusters} is a synthetically generated dynamic graph with three clusters in the \emph{causal topology}, but no pattern in the \emph{static topology}.
To generate this data set, we first constructed a random graph and generated random sequences of time-stamped edges. 
We then selectively swap the time stamps of edges such that causal walks of length two within three clusters of nodes are overrepresented, while causal walks between clusters are underrepresented.
We include a more detailed description in the appendix (code and data will be provided in a companion repository).
Apart from this synthetic data set, we use five empirical time series data sets: \textbf{student-sms} captures time-stamped SMS exchanged over four weeks between freshmen at the Technical University of Denmark \cite{sapiezynski2019interaction}.
We use the gender of participants as ground truth classes and use a maximum time difference of $\delta=40$. Since the time granularity of this data set is five minutes, this corresponds to a maximum time difference of $200$ minutes.
\textbf{high-school-2011} and \textbf{high-school-2012} capture time-stamped proximities between high-school students in two consecutive years \cite{fournet2014contact} (4 days in 2001, 7 days in 2012).
We use the gender of students as ground truth classes.
\textbf{workplace} captures time-stamped proximity interaction between employees recorded in an office building for multiple days in different years \cite{genois2015data}.
We use the department of employees as ground truth classes.
\textbf{hospital} captures time-stamped proximities between patients and healthcare workers in a hospital ward.
We use employees' roles (patient, nurse, administrative, doctor) as ground truth node classes.
All the proximity datasets were collected with a resolution of 20 seconds.
To mitigate the computational complexity of the causal walk extraction in the (undirected) proximity data sets, we coarsen the resolution by aggregating interactions to a resolution of fifteen minutes and use $\delta=4$, which corresponds to a maximum time difference of one hour.
Based on the resulting statistics of causal walks, we use the method (and code) provided in \cite{Scholtes2017} to select a higher-order De Bruijn graph model.
In \cref{table: datasetinfo} we report the $p$-value of the resulting likelihood ratio test, which is used to test the hypothesis that a first-order graph model is sufficient to explain the observed causal walk statistics, against the alternative hypothesis that a second-order De Bruijn graph model is needed.
Since all $p$-values are numerically zero, we find strong evidence for patterns that justify a second-order De Bruijn graph model for all data sets.

\begin{table}[t]
\centering \small
\resizebox{.9\textwidth}{!}{%
\begin{tabular}{l c r r r r r r r r}
\toprule
Data set & Ref & $|V|$ & $|E|$ & $|E^\mathcal{T}|$ & $p (k=2)$ & $|V^{(2)}|$ & $|E^{(2)}|$ & $\delta$ & Classes \\  
\midrule
temp-clusters & \cite{temporalclusters}  &  30      & 560  & 60000 & 0.0  &    560    &   6,789   &  1    & 3 \\
high-school-2011 & \cite{fournet2014contact} &    126   & 3042 & 28561 & 0.0  &  3042    & 17141    &    4 & 2 \\ 
high-school-2012 & \cite{fournet2014contact} &   180    & 3965 & 45047 & 0.0  &  3965    & 20614    &  4   & 2 \\ 
hospital & \cite{vanhems2013estimating}  &    75    & 2028  & 32424 & 0.0  &    2028  & 15500     & 4    & 4  \\
student-sms & \cite{sapiezynski2019interaction}     &   429    & 733 &  46138 & 0.0  &    733  &  846      & 40    & 2 \\ 
workplace   & \cite{genois2015data}  &   92    & 1431 & 9827 & 0.0  &    1431  &     7121   &    4 & 5 \\  
\bottomrule
\end{tabular}%
}
\caption{Overview of time series data and ground truth node classes used in the experiments.}
\label{table: datasetinfo}
\end{table}
Using a second-order De Bruijn graph, we compare the node classification performance of the DBGNN architecture against the following five baselines.
The first three are standard (static) graph learning techniques, namely Graph Convolutional Networks (\textbf{GCN}) \cite{Kipf2016_GCN}, \textbf{DeepWalk} \cite{Perozzi2014_DeepWalk} and \textbf{node2vec} \cite{Grover2016_node2vec}.
We further use two recently proposed temporal graph embedding techniques:
Embedding Variable Orders (\textbf{EVO}) \cite{Belth2019_evo}, is a node representation learning framework that captures non-Markovian characteristics in dynamic graphs. 
Similar to our approach, EVO uses a higher-order network to generate time-aware node representations that can be used for downstream node classification.
\textbf{HONEM} \cite{Saebi2020_HONEM} is a higher-order network embedding approach that captures non-Markovian dependencies in time series data on graphs.
This framework uses truncated SVD on a higher-order neighborhood matrix that considers the temporal order of interactions.

Addressing Q1, the results of our experiments on node classification are shown in \Cref{table:results}.
Since the classes of the empirical data sets are imbalanced, we use balanced accuracy and additionally report macro-averaged precision, recall and f1-score for a 70-30 training-test split.
We report the average performance across multiple splits. 
For DBGNN, GCN, DeepWalk, node2vec, and HONEM we performed $50$ runs.
Due to its larger computational complexity (and time constraints) we could only perform $10$ runs on EVO.
The standard deviations are included in the appendix.
We trained node2vec, EVO, and DeepWalk with $80$ walks of length $40$ per each node and a window of $10$.
We obtained the embeddings using the word2vec implementation in \cite{rehurek2011gensim}.
For EVO, we use the average as an aggregator for the higher-order representations.
To ensure the comparability of the results from GCN and DBGNN, we train both with the same number of convolutional layers with a learning rate of $0.001$ for $5000$ epochs, ELU \cite{Clevert2016FastAA} as activation function, and Adam \cite{adam} optimiser.
For DBGNN, we use SUM as aggregation function $\mathcal{F}$.
Since the data sets had no node features, we used one-hot encoding of nodes as a feature matrix (and a one-hot encoding of higher-order nodes in the initial layer of the DBGNN).
For all methods, we fix the dimensionality of the learned representations to $d = 16$, which is justified by the size of the graphs.
We manually tuned the number of hidden dimensions of the first hidden layers for GCN and DBGNN, as well as the p and q parameters of EVO and node2vec.
We report the results for the best combination of hyperparameters.

As expected, the results in \Cref{table:results} for the synthetic temporal clusters data set show that the three time-aware methods (EVO, HONEM, and DBGNN) perform considerably better than the static counterparts, which only ``see'' a random graph topology that does not allow to meaningfully assign node classes.
Both EVO and our proposed DBGNN architecture are able to perfectly classify nodes in this data set.
Interestingly, despite their good performance in the synthetic data set, the three time-aware methods show much higher variability in the empirical data sets.
We find that DBGNN shows superior performance in terms of balanced accuracy, f1-macro, and recall-macro, for all of the five empirical data sets, with a relative performance increase compared to the second best method ranging from $1.55\%$ to $28.16\%$. 
For precision-macro, DBGNN performs best in four of the five.
We attribute these results to the ability of our architecture to consider both patterns in the (static) graph topology and the causal topology, as well as to the underlying supervised approach that is due to the use of the GCN-based message passing.

To address Q2, we study visualizations of the hidden representations of higher- and first-order nodes generated by the DBGNN architecture for the synthetic temporal cluster data set, which exhibits three clear clusters in the causal topology.
We use the hidden representations $\Vec{h_v^{b}}$ generated by the bipartite layer of our DBGNN architecture, as defined in \Cref{sec:architecture}.
We compare this to the representation generated in the last message passing layer of a GCN.
\Cref{fig:latent_space_represenatation} in the appendix confirms that the DBGNN architecture learns meaningful latent space representations of nodes that incorporate temporal patterns.

\begin{table}[t]
\resizebox{.95\textwidth}{!}{%
\input{table_results_mean_only.table}
}
\caption{Results of node classification in six dynamic graphs for static graph learning techniques (DeepWalk, node2vec, GCN) and time-aware methods (HONEM, EVO) as well as the DBGNN architecture proposed in this work.}
\label{table:results}
\end{table}

\section{Conclusion}
\label{sec:conclusion}

In summary, we propose an approach to apply graph neural networks to high-resolution time series data that captures the temporal ordering of time-stamped edges in dynamic graphs.
Our method is based on a novel combination of (i) a statistical approach to infer an optimal static higher-order De Bruijn graph model for the causal topology that is due to the temporal ordering of edges, (ii) gradient-based learning in a neural network architecture that performs neural message passing in the inferred higher-order De Bruijn graph, and (iii) an additional bipartite mapping layer that maps the learnt hidden representation of higher-order nodes to the original node space.
Thanks to this approach, our architecture is able to generalize neural message passing to a \emph{static} higher-order graph model that captures the causal topology of a dynamic graph, which can considerably deviate from what we would expect based on the mere (static) topology of edges.
The results of our experiments demonstrate that the resulting architecture can considerably improve the performance of node classification in time series data, despite using message passing in a relatively simple static (augmented) graph. 
Bridging recent research on higher-order graph models in network science and deep learning in graphs \cite{eliassirad2021,benson2021higher,Lambiotte2019,Torres2021}, our work contributes to the ongoing discussion about the need for \emph{augmented message passing} schemes in data on graphs with complex characteristics \cite{velivckovic2022message}.

\paragraph{Acknowledgements}
Vincenzo Perri and Ingo Scholtes acknowledge support by the Swiss National Science Foundation, grant 176938. 
Lisi Qarkaxhija thanks Chester Tan for discussions on the manuscript.

\bibliographystyle{naturemag}

{
\small
\bibliography{library}
}

\newpage

\appendix

\section{Generation of Synthetic data with temporal clusters}

\textbf{temp-clusters} is a synthetically generated dynamic graph with a random static topology but a strong cluster structure in the causal topology.
To generate the dynamic graph, we first generate a static directed random graph with $n$ vertices and $m$ edges. 
For our experiment we chose $n=30$ and $m=560$.
We randomly assign vertices to three equally-sized, non-overlapping clusters, where $C(v)$ denotes the cluster of vertex $v$.
We then generate $N$ sequences of two randomly chosen time-stamped edges $(v_0,v_1;t)$ and $(v_1,v_2;t+1)$ that contribute to a causal walk of length two in the resulting dynamic graph.
For each vertex $v_1$ of such a causal path of length two, we randomly pick:

\begin{itemize}
	\item two time-stamped edges $(u, v_1; t_1)$ and $(v_1, w, t_1+1)$ such that $C(u)=C(v_1) \neq C(w)$
	\item two time-stamped edges $(x, v_1; t_2)$ and $(v_1, z; t_2+1)$ with $C(v_1)=C(z) \neq C(x)$
\end{itemize}

Finally, we swap the time stamps of the four time-stamped edges to $(u, v_1; t_1)$ and $(v_1, z; t_1+1)$, $(x, v_1, t_2)$, and $(v_1, w, t_2+1)$.
This swapping procedure is repeated for each vertex $v_1$ of a causal path of length two.
This simple process changes the temporal ordering of time-stamped edges, affecting neither the topology nor the frequency of time-stamped edges.
The model changes time stamps of edges (and thus causal paths) such that vertices are preferentially connected---via causal paths of length two---to other vertices in the same cluster.
This leads to a strong cluster structure in the causal topology of the dynamic graph, which (i) is neither present in the time-aggregated topology nor in the temporal activation patterns of edges, and (ii) can nevertheless be detected by higher-order methods.
A random reshuffling of timestamps destroys the cluster pattern, which confirms that it is only due to the temporal order of time-stamped edges.

\section{Latent Space Embeddings of Synthetic Example}

\begin{figure}[p]
\begin{center}
\subfigure[Latent space representation of nodes generated by De Bruijn Graph Neural Network (DBGNN) using higher-order De Bruijn graph with order $k=2$.\label{fig:latent:dbgnn}]{
\includegraphics[width=.9\textwidth]{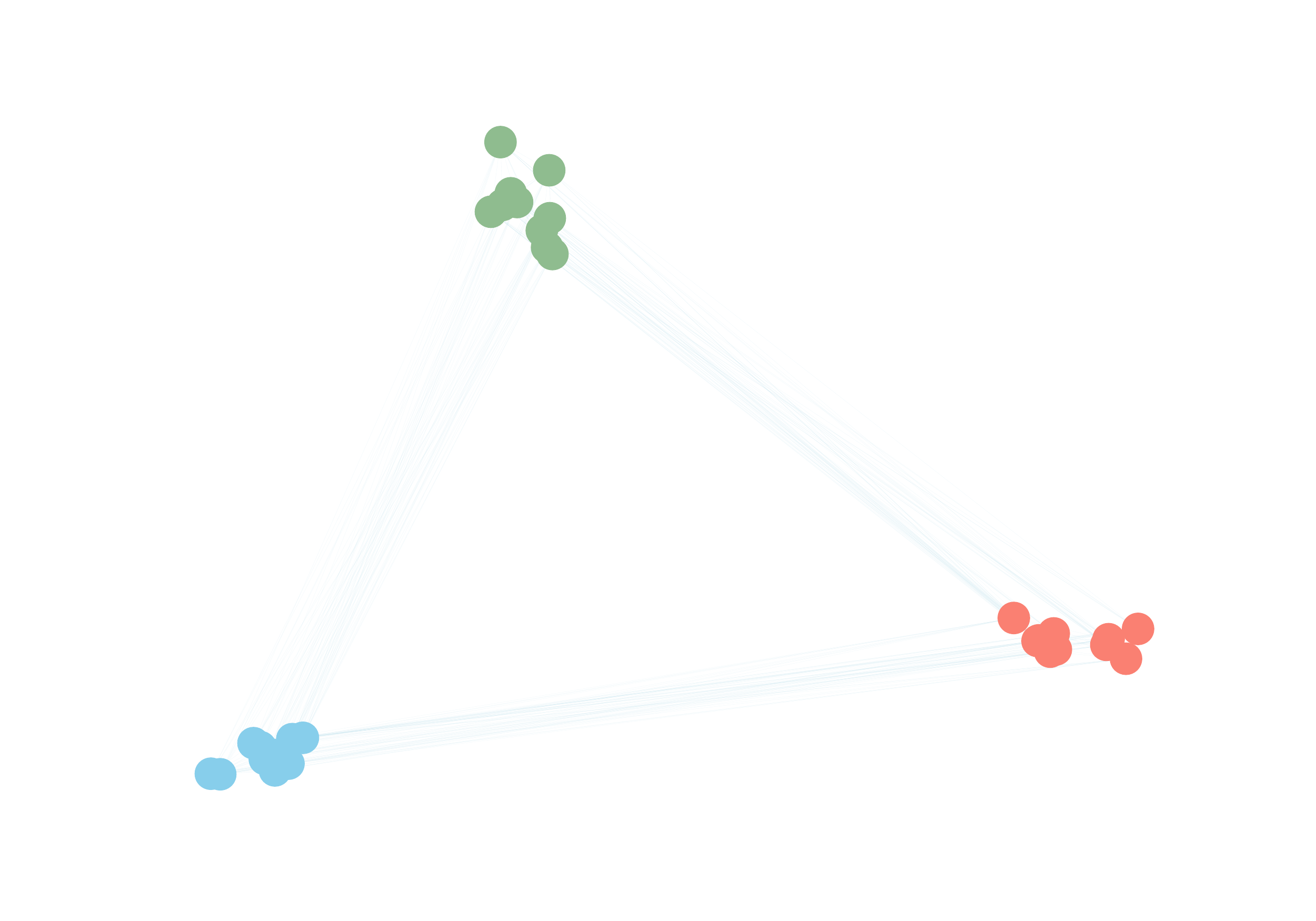}
}\\ 
\subfigure[Latent space representation of nodes generated by Graph Convolutional Network (GCN). \label{fig:latent:gcn}]{
\includegraphics[width=.9\textwidth]{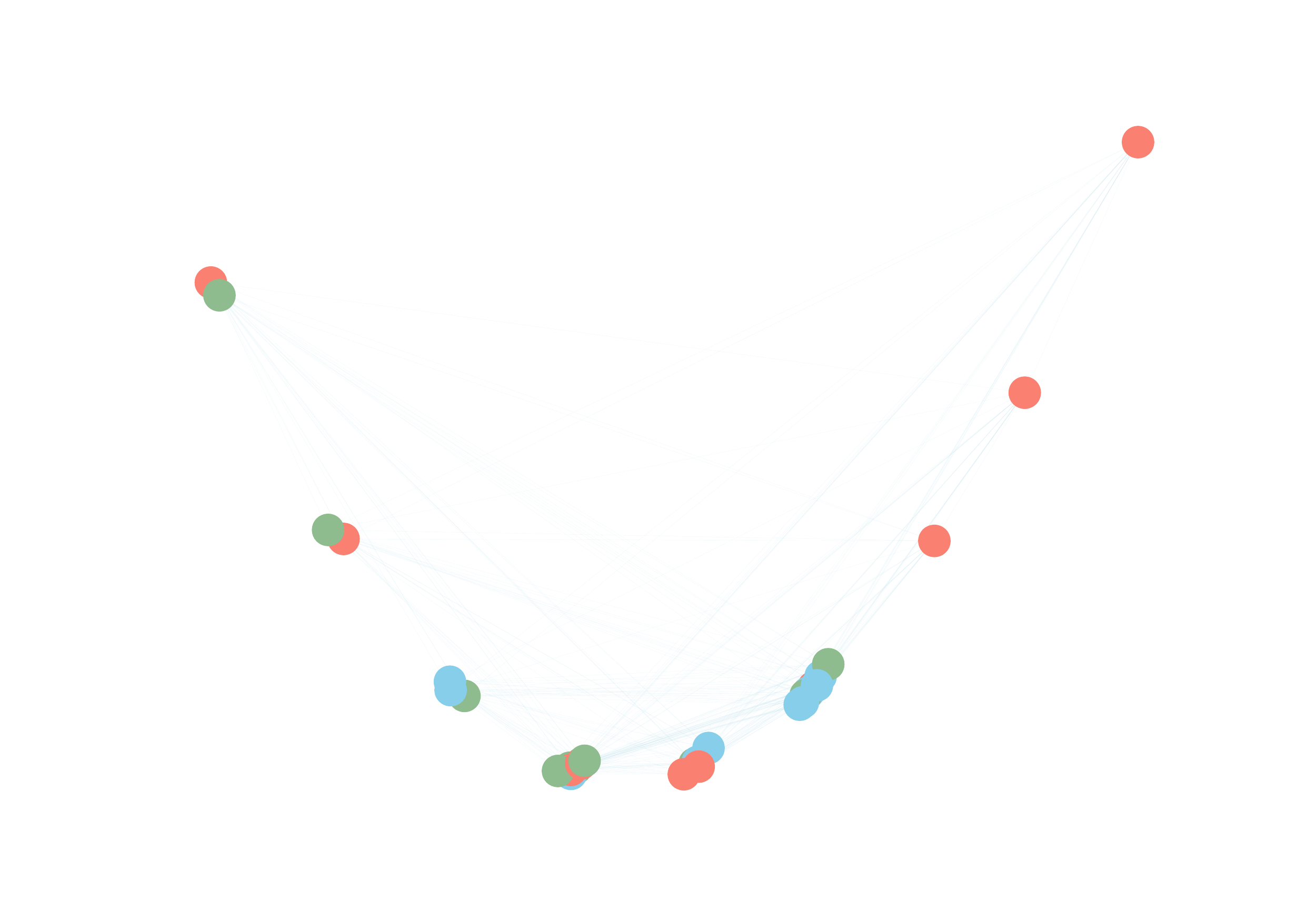}
}
\end{center}
\caption{Latent space representations of nodes in a synthetically generated dynamic graph (\textbf{temp-clusters}) with three clusters in the causal topology, where colours indicate cluster memberships. The hidden node representations learned by the DBGNN architecture capture the cluster structure in the causal topology, which is exclusively due to the temporal ordering --and not due to the topology or frequency-- of time-stamped edges.}
\label{fig:latent_space_represenatation}
\end{figure}

\Cref{fig:latent_space_represenatation} shows a latent representation of nodes in the synthetic data set temp-clusters generated by the DBGNN (a) and GCN (b) architecture.
This synthetically generated dynamic graph contains no pattern whatsoever in the (static) graph topology, which corresponds to a random graph, i.e. the topology of edges is random and all nodes have similar degrees (cf. \Cref{fig:latent:gcn}).
However, correlations in the temporal ordering of edges lead to three strong clusters in the \emph{causal topology}, i.e. there are three groups of nodes where --due to the arrow of time and the temporal ordering of edges-- pairs of nodes within the same cluster can influence each other via causal walks more frequently than pairs of nodes in different clusters.
We emphasize that the resulting pattern in the causal topology is exclusively due to the temporal ordering of edges.
The latent space embedding in \Cref{fig:latent:dbgnn} highlights the DBGNN architectures's ability to learn this pattern in the causal topology of the underlying dynamic graph, which is absent in \Cref{fig:latent:gcn}.
As expected, the different node degrees of the static graph (visible as clusters in \Cref{fig:latent:gcn}) are the only pattern captured in the hidden node representations of the GCN architecture, which is insensitive to the temporal ordering of edges.
This synthetic example confirms that DBGNNs provide a simple, static causality-aware approach for deep learning in dynamic graphs.

\newpage

\section{Standard Deviation of Classification Results}

In \Cref{table:resultsstd} we present the standard deviation of the classification results reported in \cref{table:results} across all runs for all models.

\bigskip

\begin{table}[hb]
\resizebox{\textwidth}{!}{%
\input{table_results_std_only.table}
}
\caption{Standard deviations of node classification in six dynamic graphs for static graph learning techniques (DeepWalk, node2vec, GCN) and time-aware methods (HONEM, EVO) as well as the DBGNN architecture proposed in this work.}
\label{table:resultsstd}
\end{table}

\end{document}

%% file: table_results_mean_only.table.tex
\begin{tabular}{l|l|llll}
\toprule
dataset & method                & Balanced Accuracy & F1-score-macro & Precision-macro & Recall-macro \\
\midrule
temp-clusters  
               & DeepWalk &                  32.47 &               30.39 &                32.25 &             32.47 \\
               & Node2Vec p=1 q=4 &                  35.48 &               33.02 &                34.92 &             35.48 \\
               & GCN (8,32) &                  33.52 &                12.5 &                 8.61 &             33.52 \\
               & EVO p=1 q=1 &                  \textbf{100.0} &               \textbf{100.0} &                \textbf{100.0} &             \textbf{100.0} \\
               & HONEM &                  54.94 &                53.5 &                58.16 &             54.94 \\
               & DBGNN (16,16) &                  \textbf{100.0} &               \textbf{100.0} &                \textbf{100.0} &             \textbf{100.0} \\
\hline
gain & & 0\%& 0\%& 0\%& 0\%\\
\hline
high-school-2011 

               & DeepWalk &                  55.25 &               54.02 &                60.45 &             55.25 \\
               & Node2Vec p=1 q=4 &                  56.89 &               56.29 &                60.05 &             56.89 \\
               & GCN (32,4) &                  50.06 &               40.27 &                33.99 &             50.06 \\
               & EVO p=1 q=4 &                  57.21 &               56.28 &                62.09 &             57.21 \\
               & HONEM &                  54.24 &               53.08 &                56.44 &             54.24 \\
               & DBGNN (32,8) &                   \textbf{64.4} &                \textbf{63.7} &                \textbf{65.14} &              \textbf{64.4} \\
\hline
gain & & 12.57\%& 13.16\%& 4.91\%& 12.57\%\\
\hline
high-school-2012 
               & DeepWalk &                  59.46 &                59.6 &                71.71 &             59.46 \\
               & Node2Vec p=1 q=4 &                  60.75 &               61.23 &                \textbf{72.44} &             60.75 \\
               & GCN (8,32) &                  58.03 &               56.39 &                59.16 &             58.03 \\
               & EVO p=4 q=1 &                  57.98 &                57.5 &                69.42 &             57.98 \\
               & HONEM &                  53.16 &                51.7 &                56.59 &             53.16 \\
               & DBGNN (4,8) &                   \textbf{65.8} &               \textbf{65.89} &                67.27 &              \textbf{65.8} \\
\hline
gain & & 8.31\%& 7.61\%& -7.14\%& 8.31\%\\
\hline
hospital 
               & DeepWalk &                  47.18 &               44.18 &                43.91 &             47.18 \\
               & Node2Vec p=1 q=4 &                   50.6 &               47.14 &                45.81 &              50.6 \\
               & GCN [32,32] &                  49.48 &               44.62 &                43.55 &             49.48 \\
               & EVO p=1 q=4 &                  36.34 &               36.44 &                 42.1 &             36.34 \\
               & HONEM &                  46.17 &               43.13 &                44.45 &             46.17 \\
               & DBGNN (32,16) &                  \textbf{59.04} &               \textbf{55.26} &                \textbf{58.71} &             \textbf{57.71} \\
\hline
gain & & 16.68\%& 17.23\%& 28.16\%& 14.05\%\\
\hline
student-sms 
               & DeepWalk &                  53.22 &               50.57 &                60.57 &             53.22 \\
               & Node2Vec p=1 q=4 &                  53.22 &               50.97 &                58.56 &             53.22 \\
               & GCN (4,32) &                  57.33 &               57.25 &                57.72 &             57.33 \\
               & EVO p=4 q=1 &                  52.93 &               50.66 &                57.14 &             52.93 \\
               & HONEM &                  50.43 &               44.44 &                52.91 &             50.43 \\
               & DBGNN (4,4) &                   \textbf{60.6} &               \textbf{60.89} &                \textbf{62.55} &              \textbf{60.6} \\
\hline
gain & & 5.7\%& 6.36\%& 3.27\%& 5.7\%\\
\hline
workplace 
               & DeepWalk &                  77.81 &               76.74 &                76.06 &             77.81 \\
               & Node2Vec p=1 q=4 &                   78.0 &               77.01 &                76.38 &              78.0 \\
               & GCN (32,16) &                  81.86 &               78.72 &                78.58 &             79.93 \\
               & EVO p=1 q=4 &                   77.0 &               75.68 &                75.03 &              77.0 \\
               & HONEM &                  73.26 &               72.82 &                73.73 &             73.26 \\
               & DBGNN (32,8) &                  \textbf{83.13} &               \textbf{81.06} &                \textbf{81.52} &             \textbf{81.75} \\
\hline
gain & & 1.55\%& 2.97\%& 3.74\%& 2.28\%\\
\bottomrule
\end{tabular}

%% file: table_results_std_only.table.tex
\begin{tabular}{l|l|llll}
\toprule
dataset & method &  Balanced Accuracy & F1-score-macro & Precision-macro & Recall-macro \\
\midrule
   temp-clusters & DeepWalk &                  15.38 &               15.04 &                18.03 &             15.38 \\
               & Node2Vec p=1 q=4 &                  17.12 &               16.88 &                20.24 &             17.12 \\
               & GCN (8,32) &                    7.3 &                7.69 &                 8.04 &               7.3 \\
                & EVO p=1 q=1 &                    0.0 &                 0.0 &                  0.0 &               0.0 \\

               & HONEM &                  16.27 &               16.71 &                19.61 &             16.27 \\
                & DBGNN (16,16) &                    0.0 &                 0.0 &                  0.0 &               0.0 \\

\hline
high-school-2011  & DeepWalk &                   5.83 &                7.22 &                12.79 &              5.83 \\
               & Node2Vec1.04.0 &                   6.34 &                7.58 &                 9.44 &              6.34 \\
               & GCN (32,4) &                   0.89 &                 3.1 &                 4.83 &              0.89 \\
               & EVO p=1 q=4 &                   5.72 &                7.65 &                 9.33 &              5.72 \\

               & HONEM &                   5.72 &                6.93 &                10.07 &              5.72 \\
            & DBGNN (32,8) &                    7.0 &                7.42 &                  7.8 &               7.0 \\
\hline
high-school-2012
               & DeepWalk &                   4.97 &                6.52 &                 11.0 &              4.97 \\
               & Node2Vec p=1 q=4 &                   5.27 &                 6.8 &                11.29 &              5.27 \\
               & GCN (8,32) &                   6.87 &                9.49 &                13.58 &              6.87 \\
               & EVO p=4 q=1 &                   4.14 &                6.07 &                 9.96 &              4.14 \\

               & HONEM &                   4.59 &                5.89 &                 9.12 &              4.59 \\
            & DBGNN (4,8) &                   6.59 &                6.62 &                 7.07 &              6.59 \\
\hline
hospital
               & DeepWalk &                   7.64 &                 6.9 &                 7.51 &              7.64 \\
               & Node2Vec p=1 q=4 &                   6.79 &                6.46 &                 6.95 &              6.79 \\
               & GCN (32,32) &                  11.06 &                12.0 &                13.58 &             11.06 \\
               & EVO p=1 q=4 &                   9.31 &               11.34 &                16.31 &              9.31 \\
               & HONEM &                   8.51 &                7.78 &                 8.25 &              8.51 \\
            & DBGNN (32,16) &                  13.09 &               12.54 &                15.02 &             12.65 \\
\hline
student-sms 
               & DeepWalk &                   2.72 &                4.45 &                10.05 &              2.72 \\
               & Node2Vec p=1 q=4 &                   3.29 &                4.93 &                 9.13 &              3.29 \\
               & GCN (4,32) &                   3.59 &                3.65 &                 3.91 &              3.59 \\
               & EVO p=4 q=1 &                   3.38 &                5.14 &                 7.89 &              3.38 \\
               & HONEM &                   1.29 &                2.31 &                 15.0 &              1.29 \\
            & DBGNN (4,4) &                   4.28 &                4.47 &                 4.56 &              4.28 \\
\hline
workplace 
               & DeepWalk &                   2.23 &                1.85 &                 1.48 &              2.23 \\
               & Node2Vec p=1 q=4 &                    3.3 &                3.11 &                 2.95 &               3.3 \\
               & GCN (32,16) &                   8.67 &                 8.6 &                 9.61 &              8.26 \\
               & EVO p=1 q=4 &                   3.12 &                2.36 &                 1.65 &              3.12 \\
               & HONEM &                   6.27 &                5.17 &                 4.34 &              6.27 \\
            & DBGNN (32,8) &                   9.67 &                9.76 &                10.26 &              9.65 \\
\bottomrule
\end{tabular}